%% file: main.tex
\documentclass[10pt,twocolumn,letterpaper]{article}

\usepackage{cvpr}              

\input{preamble}

\definecolor{cvprblue}{rgb}{0.21,0.49,0.74}
\usepackage[pagebackref,breaklinks,colorlinks,citecolor=cvprblue]{hyperref}

\def\paperID{*****} 
\def\confName{CVPR}
\def\confYear{2024}

\title{1st Place Solution to the 8th HANDS Workshop Challenge - ARCTIC Track: 3DGS-based Bimanual Category-agnostic Interaction Reconstruction}

\author{Jeongwan On \quad Kyeonghwan Gwak \quad Gunyoung Kang \quad Hyein Hwang \\ \quad Soohyun Hwang \quad Junuk Cha \quad Jaewook Han \quad Seungryul Baek \\
UNIST
}

\usepackage{float}

\begin{document}
\maketitle

\input{sec/0_abstract}
\input{sec/1_intro}
\input{sec/3_method}
\input{sec/4_experiments}
\input{sec/5_conclusion}

{
    \small
    \bibliographystyle{ieeenat_fullname}
    \bibliography{main}
}


\end{document}

%% file: preamble.tex
\usepackage[dvipsnames]{xcolor}


%% file: sec/0_abstract.tex
\begin{abstract}

This report describes our 1st place solution to the 8th HANDS workshop challenge (ARCTIC \cite{fan2023arctic} track) in conjunction with ECCV 2024. In this challenge, we address the task of \textbf{bimanual} category-agnostic hand-object interaction reconstruction, which aims to generate 3D reconstructions of both hands and the object from a monocular video, without relying on predefined templates. This task is particularly challenging due to the significant occlusion and dynamic contact between the hands and the object during bimanual manipulation. We worked to resolve these issues by introducing a mask loss and a 3D contact loss, respectively. Moreover, we applied 3D Gaussian Splatting (3DGS) to this task. As a result, our method achieved a value of \textbf{38.69} in the main metric, CD$_h$, on the ARCTIC test set.

\end{abstract}

%% file: sec/1_intro.tex
\section{Introduction}
\label{sec:intro}


\par Most hand-object interaction reconstruction methods~\cite{fan2023arctic,Faneccv2024,juchacvpr2024,cho2023transformer,armagan2020measuring,baek2020weakly,garcia2018first} rely on predefined templates for hand and object. While template-based approaches effectively leverage prior knowledge to reconstruct high-dimensional information from limited input, they come with notable limitations. First, these methods are typically restricted to specific object categories, making them difficult to apply in real-world, in-the-wild scenarios where the diversity of objects is vast. Second, they struggle with capturing fine-grained details, resulting in less accurate reconstructions of complicated hand-object interactions.

\begin{figure}[h]
  \centering
   \includegraphics[width=\linewidth]{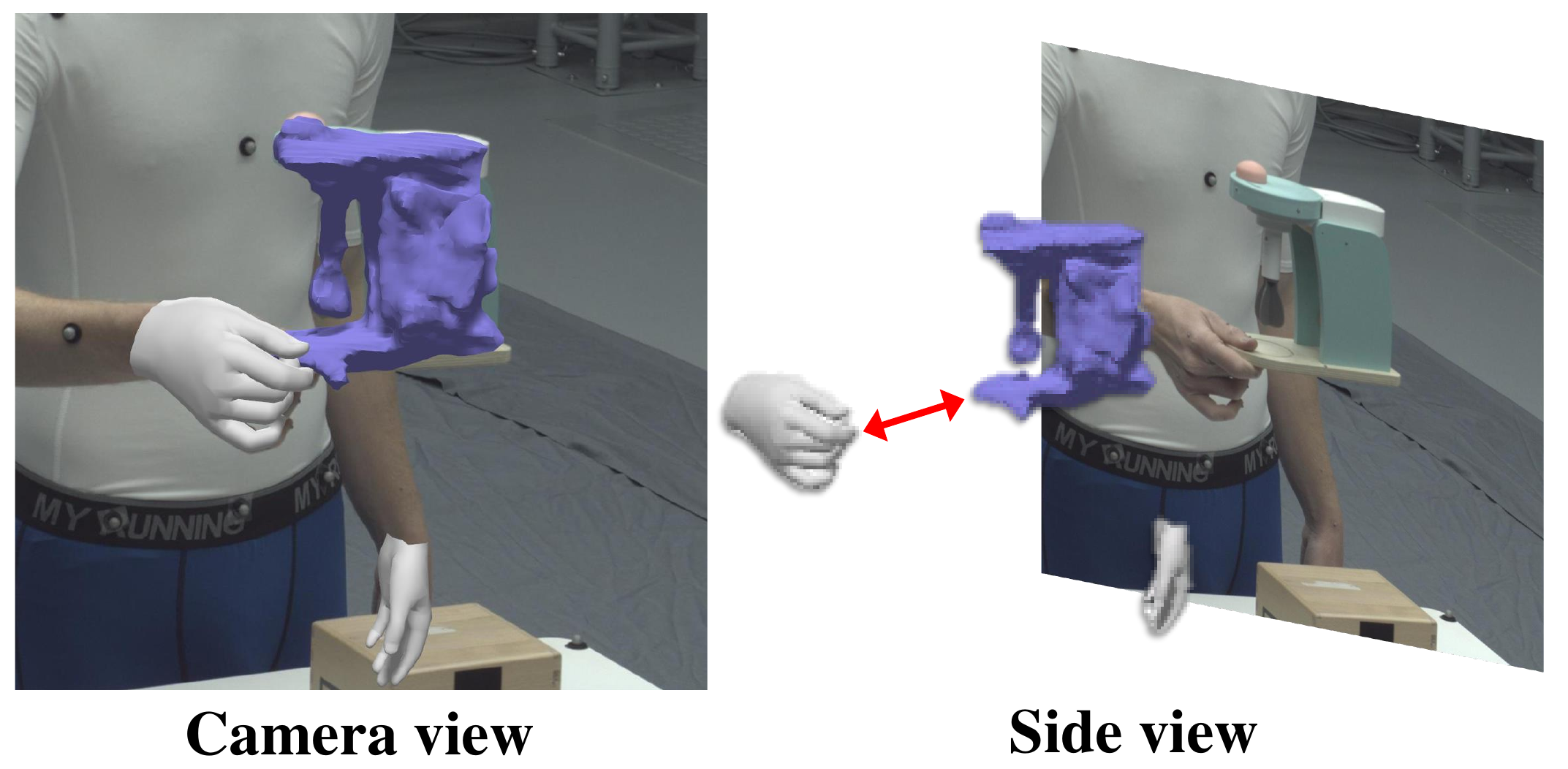}

   \caption{\textbf{Limitation of the HOLD baseline}. From the original camera viewpoint, HOLD performs well on 2D contact reconstruction. However, it performs poorly in 3D contact reconstruction when seen from different camera viewpoints. As a result, it fails to accurately estimate the relative distance between the hand and the object, which worsens the main metric, CD$_h$.}
   \label{fig:fig1}
\end{figure}
\par To address these challenges, HOLD \cite{fan2024hold} introduced a category-agnostic approach to hand-object interaction reconstruction, offering a promising solution to overcome the constraints of template-based methods. However, HOLD  is also limited to interactions involving a single hand and primarily addressed scenarios where the hand and the object were almost always in contact. As a result, the two-hand manipulation settings of HOLD showed a significant error with a CD$_h$ \cite{fan2024hold}, the hand-relative Chamfer distance of 114.73, contrast to its excellent performance in single-hand settings, where it achieved a CD$_h$ of 11.3 on the HO3D dataset. Figure \ref{fig:fig1} shows the poor 3D contact reconstruction quality of the fully trained HOLD baseline.

\begin{figure*}[t]
  \centering
   \includegraphics[width=1.0\linewidth]{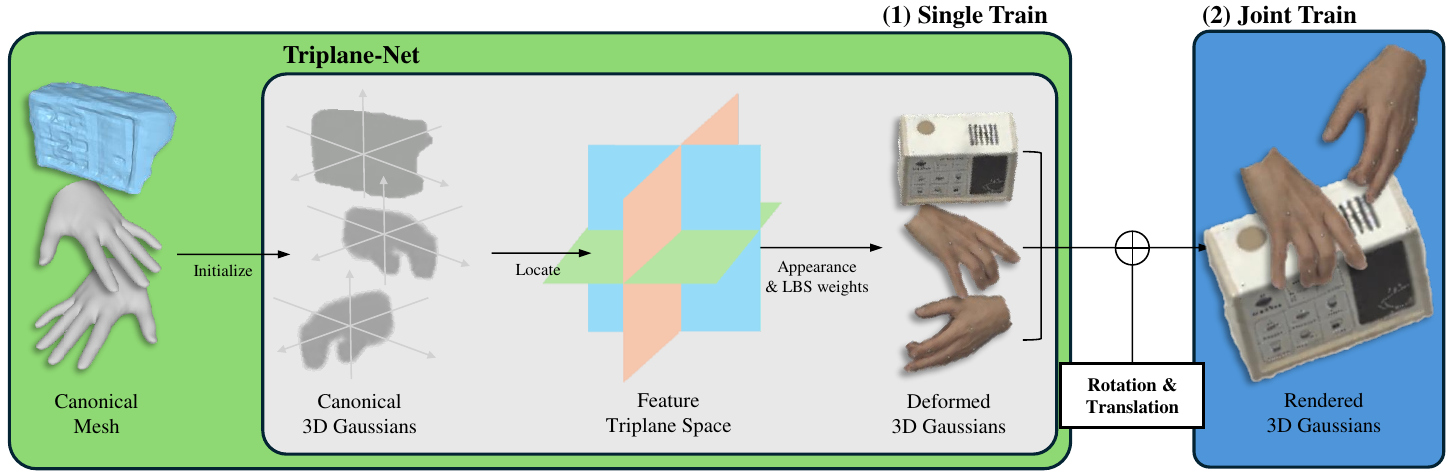}

   \caption{Our method is composed of `Single Train' and `Joint Train' stages. In `Single Train' stage, appearances and geometries for left and right hands and the object are reconstructed by fitting 3D Gaussian splats on each agent. In `Joint Train' stage, we further consider contacts between the hands and the object and refine obtained Gaussian splats.}
   \label{fig:fig2}
\end{figure*}

\par In this work, we push the boundaries of existing methods to address the challenging problem of category-agnostic reconstruction of hands and objects in bimanual settings. This task involves significant occlusion and dynamic contact between the hands and the object, making it particularly difficult to solve. To tackle these issues, we introduce a mask loss and a 3D contact loss to specifically manage occlusion and contact dynamics, respectively. Additionally, we incorporate 3D Gaussian Splatting (3DGS) \cite{kerbl3Dgaussians}, a technique that has shown great success across multiple domains, to further improve the reconstruction performance.

%% file: sec/3_method.tex
\section{Method}
\label{sec:method}

In this section, we introduce our method,
which is illustrated in Fig. \ref{fig:fig2}. Our method can split into two stages: (1) single train and (2) joint train. Stage (1) is again divided into two stages that fit (1-1) Hand Gaussian Splats and (1-2) Object Gaussian Splats, respectively. Based on \cite{hugs}, we use Triplane-Net as the baseline of our framework. We introduce more details in the following subsection.


\subsection{Triplane-Net}
According to \cite{hugs}, we first extract canonical 3D meshes via off-the-shelf model \cite{fan2024hold}, and Triplane-Net takes it as the input. Then, the extracted 3D meshes initialize 3D Gaussians and Triplane-Net locates it to the feature triplane space. Afterwards, Triplane-Net estimates the deformation, appearance, and geometry of 3D Gaussians (e.g., their LBS weight, color, rotation, and scale), and deforms Gaussians as $\Tilde{G} = \mathcal{T} (G)$ where $\mathcal{T} (\cdot)$ is Triplane-Net, $G$ is extracted canonical 3D mesh, and $\Tilde{G}$ is deformed 3D canonical mesh.

\subsection{Single Train}

{\bf Hand Gaussian Splats.} Given an input video with the length of $T$, we first extract a hand pose parameter in the $t$-th frame for left and right hand ($x \in \{ l, \; r \}$ denotes the type of hand) that includes global orientation $\Phi_{x}^t \in \mathbb{R}^{3}$, translation $\Gamma_{x}^t \in \mathbb{R}^{3}$, pose parameter $\theta_{x}^t \in \mathbb{R}^{45}$, and shape parameter $\beta_{x}^t \in \mathbb{R}^{10}$, over all frames using the off-the-shelf model \cite{fan2024hold}. Then canonical 3D hand mesh $G_x$ is obtained through the following process:
\begin{equation}
    G^{t}_x = \mathcal{M}(\theta_{x}^t, \beta_{x}^t),
  \label{eq:mano_layer}
\end{equation}

\noindent where $\mathcal{M}$ is MANO Layer~\cite{MANO:SIGGRAPHASIA:2017,baek2019pushing}. Subsequently, the deformed Gaussians $\Tilde{G}_x$ are obtained by $\mathcal{T} (G_x)$, and they are located in the camera space using the formula as follows:
\begin{equation}
    {}^{c}\Tilde{G}^{t}_{x} = \Tilde{G}^{t}_x \times \Phi^{t}_{x} + \Gamma^{t}_{x},
  \label{eq:hand_can_to_cam}
\end{equation}

\noindent where ${}^{c}\Tilde{G}_x$ is deformed camera-coordinated Gaussians for each hand.

\noindent{\bf Object Gaussian Splats.} 
\noindent Similar to Hand Gaussian Splats, we also estimate the object parameter for the $t$-th frame, including canonical 3D object mesh $G^{t}_o$ and their rotations $\Phi^{t}_o$, and translations $\Gamma^{t}_o$ using the off-the-shelf model \cite{fan2024hold}. Then, the estimated mesh is fed into the Triplane-Net, and located in the camera space using the formula as follows:
\begin{equation}
    {}^{c}\Tilde{G}^{t}_{o} = \Tilde{G}^{t}_o \times \Phi^{t}_{o} + \Gamma^{t}_{o}.
  \label{eq:obj_can_to_cam}
\end{equation}

\noindent{\bf Optimization.} We optimize the center translation of each Gaussian $\mu$, parameters of Triplane-Net $w$, and the parameter $P_{x}=\{\Phi_x, \Gamma_x, \theta_x, \beta_x \}$, where $x \in \{ l, r, o \}$ and $\theta_o = \beta_o = \varnothing$. we define the loss $\mathcal{L}_{basic}$ by combining SSIM, VGG and LBS losses of~\cite{hugs} as follows:
\begin{equation}    \mathcal{L}_{basic} = \sum_{x\in\{l,r,o\}}\lambda_1 \mathcal{L}^{x}_{ssim} + \lambda_2 \mathcal{L}^{x}_{vgg} + \lambda_3 \mathcal{L}^{x}_{LBS},  \label{eq:obj_param}
\end{equation}
\noindent where $\lambda_1 = 0.2, \lambda_2 = 1.0, \lambda_3 = 1000$. Additionally, to prevent outlier Gaussians, we employ the mask loss $\mathcal{L}_{mask}$. Specifically, we first obtain the masks of hands and an object via \cite{ravi2024sam2}. Then, we enforce that Gaussians are generated inside the mask, by using the formula as follows:
\begin{eqnarray}
\mathcal{L}_{mask} =  \sum_{t=1}^{T} {\lambda_4 \| {m}^t_x \odot (\mathbb{I}^t_{pred} -  \mathbb{I}^t_{gt}) \|^2_2   + \lambda_5  (1 - {\Bar{m}}^t) \odot \mathbb{I}^t_{pred} }    \label{eq:mask_loss}
\end{eqnarray}
where $x \in \{ l,r,o \}$ denotes left, right hands or an object, and $\odot$ denotes the element (pixel)-wise product. $m_x^t$ is obtained mask at the $t$-th frame, and $\mathbb{I}_{gt}^t$ and $\mathbb{I}_{pred}^t$ are ground-truth and rendered $t$-th frame, respectively. 
$\Bar{m}^t=\sum_x {m_x}^t$ is the merged foreground mask for two hands and an object.


We also use additional regularization term following~\cite{moon2024exavatar}, to improve the rendering quality:
\begin{equation}    \mathcal{L}_{render} = \sum_{x\in\{l,r,o\}}\lambda_6 \mathcal{L}^{x}_{color} + \lambda_7 \mathcal{L}^{x}_{scale},
  \label{eq:reg_term}
\end{equation}

\noindent where $\lambda_6 = 0.1, \lambda_7 = 100.0$. Finally, we find the $\mu$, $w$ and $P_x^t$ for all frame $t\in[1,T]$ and for left, right hands and an object that minimizes the objective defined as follows:
\begin{equation}
    \underset{ 
        \begin{smallmatrix}
            \\ \mu, w, \\
            \{ \{ P^t_x \}^{T}_{t=1} \}_{x \in \{l,r,o\}}
        \end{smallmatrix}
    }{\min} \mathcal{L}_{basic} + \mathcal{L}_{mask} + \mathcal{L}_{render}.  \label{eq:single_optim}
\end{equation}

\noindent We optimize Eq.~\ref{eq:single_optim} for 45K iteration with 1 number of NVIDIA A6000 GPU. We use gradually decreasing learning late from $1.6\times10^{-4}$ to $1.6\times10^{-6}$ for $\mu$, and use $1.0\times10^{-4}$ for other parameters.


\subsection{Joint Train}

{\bf Hand \& Object Gaussian Splats.} Most existing methods \cite{fan2024hold, pokhariya2024manus} for bimanual category-agnostic hand-object interaction reconstruction tasks do not consider contact between hand and object in 3D space or only consider it for very limited cases. For that reason, we propose a simple contact regularization term $\mathcal{L}_{contact}$ to encourage hand-object Gaussians to be well-contacted during the optimization. Towards the goal, we additionally optimize the hand translation $\Gamma_x$, where $x \in \{ l, r \}$ by employing the 3D contact regularization term $\mathcal{L}_{contact}$ to tightly contact hand and object meshes in the 3D space:
\begin{equation}    \mathcal{L}_{contact} = \lambda_8 \sum^{T}_{t=1} \sum_{x \in \{l, r \}}{ || \Gamma^t_o - \Gamma^t_x ||_2 }.
  \label{eq:contact_loss}
\end{equation}

\noindent To prevent hand Gaussians too much follow the object translation, we set the $\lambda_8$ as a small number 1.0.

Finally, we find the $\Gamma_x^t$ for $t\in[1,T]$ and $x\in\{l, r\}$ by minimizing the objective, which is defined as follows:
\begin{equation}
    \underset{ 
        \{ \{ \Gamma^t_x \}^{T}_{t=1} \}_{x \in \{l,r\}}
    }{\min} \mathcal{L}_{basic} + \mathcal{L}_{mask} + \mathcal{L}_{contact}.
  \label{eq:joint_optim}
\end{equation}

%% file: sec/4_experiments.tex
\section{Experiments}
\label{sec:experiments}

\noindent {\bf Setup.} We used 9 objects from the ARCTIC~\cite{fan2023arctic} dataset as the training dataset, with subject 3 and camera index 1. In particular, we used the action of grabbing objects and selected the 300 frames in which the hand and the object were most clearly visible. We use CD$_h$~\cite{fan2024hold} as the main metric. To further evaluate the object reconstruction, we also used CD and the F10 metric of~\cite{fan2024hold}.

\begin{figure}[htb!]
  \centering
   \includegraphics[width=0.98\linewidth]{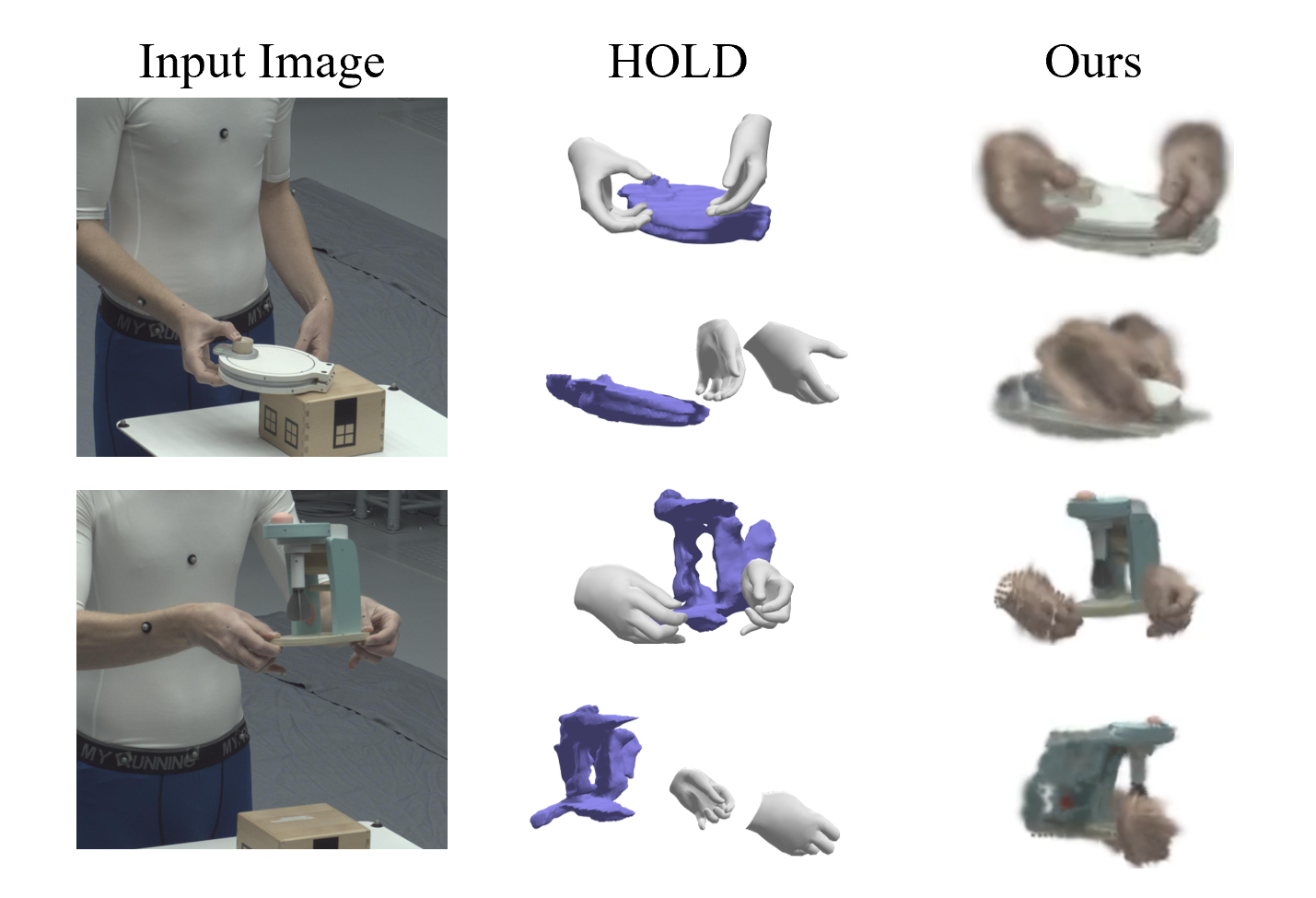}

   \caption{\textbf{Qualitative results}. For each example, the first row visualizes a result in the camera view and the second row visualizes a result in the side view. We can observe that our method provides better alignment between the hand and the object in the side view.}
   \label{fig:fig3}
\end{figure}

\noindent {\bf Results.} Table~\ref{table:table1} shows the quantitative results comparing our method with other methods submitted to the challenge server. Additionally, Fig.~\ref{fig:fig3} shows the qualitative results of our method compared with the HOLD baseline.

\begin{figure}[htb!]
  \centering
   \includegraphics[width=\linewidth]{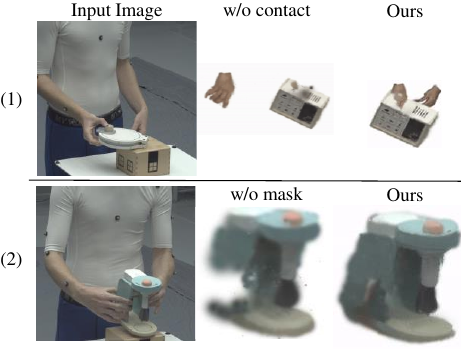}

   \caption{\textbf{Ablation study}. (1) $L_{contact}$ encourages contact between the hand and the object in the 3D space. (2)When $m$ is used instead of $\bar{m}$, the Gaussian is destroyed due to self-occlusion.}
   \label{fig:fig4}
\end{figure}

{
    \begin{table}[t]
    \centering
    \begin{tabular}{ccccc} 
    \toprule
         & CD$_h \; [cm^2] \downarrow$    & CD$ \; [cm^2] \downarrow$  & F10$ \; [\%] \uparrow$  \\ 
    \hline
    HOLD \cite{fan2024hold} & 114.73 & 2.07  & 63.92  \\
    ACE \cite{ace}  & 100.33 & 2.03  & 69.4  \\
    Ours & \textbf{38.69}  & \textbf{1.36}  & \textbf{81.78}  \\
    \bottomrule
    \end{tabular}

    \caption{\textbf{Quantitative results}. Our method achieved the best object reconstruction performance.}
    \label{table:table1}
    \end{table}
}

\noindent \textbf{Ablation study.} Figure~\ref{fig:fig4} shows our ablation examples.  when 3D contact loss is not applied, contact between the hand and the object in 3D space is become unnatural (1st row). We also observed that the reconstructed mesh breaks due to self-occlusion when mask loss is not used (2nd row).

%% file: sec/5_conclusion.tex
\section{Conclusion}
\label{sec:conclusion}

In this document, we involved our method for reconstructing hand and object meshes using 3D Gaussian splats. Our method ranked as the 1st in the 8th HANDS Workshop Challenge held with ECCV'24 -- ARCTIC track.





%% file: main.bbl
\begin{thebibliography}{16}
\providecommand{\natexlab}[1]{#1}
\providecommand{\url}[1]{\texttt{#1}}
\expandafter\ifx\csname urlstyle\endcsname\relax
  \providecommand{\doi}[1]{doi: #1}\else
  \providecommand{\doi}{doi: \begingroup \urlstyle{rm}\Url}\fi

\bibitem[Armagan et~al.(2020)Armagan, Garcia-Hernando, Baek, Hampali, Rad, Zhang, Xie, Chen, Zhang, Xiong, Xiao, Cao, Yuan, Ren, Huang, haifeng sun, Hrúz, Kanis, Krňoul, Wan, Li, Lee, Yang, Yao, Liu, Spurr, Molchanov, Iqbal, Weinzaepfel, Brégier, Rogez, Lepetit, and Kim]{armagan2020measuring}
Anil Armagan, Guillermo Garcia-Hernando, Seungryul Baek, Shreyas Hampali, Mahdi Rad, Zhaohui Zhang, Shipeng Xie, Neo Chen, Boshen Zhang, Fu Xiong, Yang Xiao, Zhiguo Cao, Junsong Yuan, Pengfei Ren, Weiting Huang, haifeng sun, Marek Hrúz, Jakub Kanis, Zdeněk Krňoul, Qingfu Wan, Shile Li, Dongheui Lee, Linlin Yang, Angela Yao, Yun-Hui Liu, Adrian Spurr, Pavlo Molchanov, Umar Iqbal, Philippe Weinzaepfel, Romain Brégier, Grégory Rogez, Vincent Lepetit, and Tae-Kyun Kim.
\newblock {Measuring generalisation to unseen viewpoints, articulations, shapes and objects for 3D hand pose estimation under hand-object interaction}.
\newblock In \emph{ECCV}, 2020.

\bibitem[Baek et~al.(2019)Baek, Kim, and Kim]{baek2019pushing}
Seungryul Baek, Kwang~In Kim, and Tae-Kyun Kim.
\newblock Pushing the envelope for rgb-based dense 3d hand pose estimation via neural rendering.
\newblock In \emph{CVPR}, 2019.

\bibitem[Baek et~al.(2020)Baek, Kim, and Kim]{baek2020weakly}
Seungryul Baek, Kwang~In Kim, and Tae-Kyun Kim.
\newblock Weakly-supervised domain adaptation via gan and mesh model for estimating 3d hand poses interacting objects.
\newblock In \emph{CVPR}, 2020.

\bibitem[Cha et~al.(2024)Cha, Kim, Yoon, and Baek]{juchacvpr2024}
Junuk Cha, Jihyeon Kim, Jae~Shin Yoon, and Seungryul Baek.
\newblock Text2hoi: Text-guided 3d motion generation for hand-object interaction.
\newblock In \emph{CVPR}, 2024.

\bibitem[Cho et~al.(2023)Cho, Kim, Kim, Lee, Ismayilzada, and Baek]{cho2023transformer}
Hoseong Cho, Chanwoo Kim, Jihyeon Kim, Seongyeong Lee, Elkhan Ismayilzada, and Seungryul Baek.
\newblock Transformer-based unified recognition of two hands manipulating objects.
\newblock In \emph{CVPR}, 2023.

\bibitem[Fan et~al.(2023)Fan, Taheri, Tzionas, Kocabas, Kaufmann, Black, and Hilliges]{fan2023arctic}
Zicong Fan, Omid Taheri, Dimitrios Tzionas, Muhammed Kocabas, Manuel Kaufmann, Michael~J. Black, and Otmar Hilliges.
\newblock {ARCTIC}: A dataset for dexterous bimanual hand-object manipulation.
\newblock In \emph{CVPR}, 2023.

\bibitem[Fan et~al.(2024{\natexlab{a}})Fan, Ohkawa, Yang, Lin, Zhou, Zhou, Liang, Gao, Zhang, Zhang, Li, Liu, Lu, Zeid, Leibe, On, Baek, Prakash11, Gupta, He, Sato, Hilliges, Chang, and Yao]{Faneccv2024}
Zicong Fan, Takehiko Ohkawa, Linlin Yang, Nie Lin, Zhishan Zhou, Shihao Zhou, Jiajun Liang, Zhong Gao, Xuanyang Zhang, Xue Zhang, Fei Li, Zheng Liu, Feng Lu, Karim~Abou Zeid, Bastian Leibe, Jeongwan On, Seungryul Baek, Aditya Prakash11, Saurabh Gupta, Kun He, Yoichi Sato, Otmar Hilliges, Hyung~Jin Chang, and Angela Yao.
\newblock Benchmarks and challenges in pose estimation for egocentric hand interactions with objects.
\newblock In \emph{ECCV}, 2024{\natexlab{a}}.

\bibitem[Fan et~al.(2024{\natexlab{b}})Fan, Parelli, Kadoglou, Kocabas, Chen, Black, and Hilliges]{fan2024hold}
Zicong Fan, Maria Parelli, Maria~Eleni Kadoglou, Muhammed Kocabas, Xu Chen, Michael~J Black, and Otmar Hilliges.
\newblock {HOLD}: Category-agnostic 3d reconstruction of interacting hands and objects from video.
\newblock In \emph{CVPR}, 2024{\natexlab{b}}.

\bibitem[Garcia-Hernando et~al.(2018)Garcia-Hernando, Yuan, Baek, and Kim]{garcia2018first}
Guillermo Garcia-Hernando, Shanxin Yuan, Seungryul Baek, and Tae-Kyun Kim.
\newblock First-person hand action benchmark with rgb-d videos and 3d hand pose annotations.
\newblock In \emph{CVPR}, 2018.

\bibitem[Kerbl et~al.(2023)Kerbl, Kopanas, Leimk{\"u}hler, and Drettakis]{kerbl3Dgaussians}
Bernhard Kerbl, Georgios Kopanas, Thomas Leimk{\"u}hler, and George Drettakis.
\newblock 3d gaussian splatting for real-time radiance field rendering.
\newblock \emph{ACM ToG}, 2023.

\bibitem[Kocabas et~al.(2024)Kocabas, Chang, Gabriel, Tuzel, and Ranjan]{hugs}
Muhammed Kocabas, Rick Chang, James Gabriel, Oncel Tuzel, and Anurag Ranjan.
\newblock Hugs: Human gaussian splats.
\newblock In \emph{CVPR}, 2024.

\bibitem[Moon et~al.(2024)Moon, Shiratori, and Saito]{moon2024exavatar}
Gyeongsik Moon, Takaaki Shiratori, and Shunsuke Saito.
\newblock Expressive whole-body {3D} gaussian avatar.
\newblock In \emph{ECCV}, 2024.

\bibitem[Pokhariya et~al.(2024)Pokhariya, Shah, Xing, Li, Chen, Sharma, and Sridhar]{pokhariya2024manus}
Chandradeep Pokhariya, Ishaan~Nikhil Shah, Angela Xing, Zekun Li, Kefan Chen, Avinash Sharma, and Srinath Sridhar.
\newblock Manus: Markerless grasp capture using articulated 3d gaussians.
\newblock In \emph{CVPR}, 2024.

\bibitem[Ravi et~al.(2024)Ravi, Gabeur, Hu, Hu, Ryali, Ma, Khedr, R{\"a}dle, Rolland, Gustafson, Mintun, Pan, Alwala, Carion, Wu, Girshick, Doll{\'a}r, and Feichtenhofer]{ravi2024sam2}
Nikhila Ravi, Valentin Gabeur, Yuan-Ting Hu, Ronghang Hu, Chaitanya Ryali, Tengyu Ma, Haitham Khedr, Roman R{\"a}dle, Chloe Rolland, Laura Gustafson, Eric Mintun, Junting Pan, Kalyan~Vasudev Alwala, Nicolas Carion, Chao-Yuan Wu, Ross Girshick, Piotr Doll{\'a}r, and Christoph Feichtenhofer.
\newblock Sam 2: Segment anything in images and videos.
\newblock \emph{ArXiv Preprint:2408.00714}, 2024.

\bibitem[Romero et~al.(2017)Romero, Tzionas, and Black]{MANO:SIGGRAPHASIA:2017}
Javier Romero, Dimitrios Tzionas, and Michael~J. Black.
\newblock Embodied hands: Modeling and capturing hands and bodies together.
\newblock \emph{ACM ToG}, 2017.

\bibitem[Xu et~al.(2024)Xu, Liu, Cui, and Yan]{ace}
Congsheng Xu, Yitian Liu, Yi Cui, and Yichao Yan.
\newblock Ace, 2024.

\end{thebibliography}
